\documentclass{article}

\usepackage{PRIMEarxiv}

\usepackage[utf8]{inputenc} 
\usepackage[T1]{fontenc}    
\usepackage{hyperref}       
\usepackage{url}            
\usepackage{booktabs}       
\usepackage{amsfonts}       
\usepackage{nicefrac}       
\usepackage{microtype}      
\usepackage{lipsum}
\usepackage{fancyhdr}       
\usepackage{graphicx}       
\usepackage{amsmath}
\usepackage{tikz}
\usetikzlibrary{arrows.meta, positioning, shapes.misc, shapes.geometric}
\usetikzlibrary{calc, backgrounds, fit}

\graphicspath{{media/}}     

\title{BIONIX: A Wireless, Low-Cost Prosthetic Arm with Dual-Signal EEG and EMG Control}

\author{
  Pranesh Sathish Kumar \\
  Alliance Academy for Innovation \\
  Cumming, GA\\
  \texttt{\ praneshsathishk@gmail.com} \\
  \texttt{\ pkumar418@gatech.edu} \\
}

\begin{document}
\maketitle

\begin{abstract}
Affordable upper-limb prostheses often lack intuitive control systems, limiting functionality and accessibility for amputees in low-resource settings. This project presents a low-cost, dual-mode neuro-muscular control system integrating electroencephalography (EEG) and electromyography (EMG) to enable real-time, multi-degree-of-freedom control of a prosthetic arm. EEG signals are acquired using the NeuroSky MindWave Mobile 2 and transmitted via ThinkGear Bluetooth packets to an ESP32 microcontroller running a lightweight classification model. The model was trained on 1500 seconds of recorded EEG data using a 6-frame sliding window with low-pass filtering, excluding poor-signal samples and using a 70/20/10 training--validation--test split. The classifier detects strong blink events, which toggle the hand between open and closed states. EMG signals are acquired using a MyoWare 2.0 sensor and SparkFun wireless shield and transmitted to a second ESP32, which performs threshold-based detection. Three activation bands (rest: 0--T1; extension: T1--T2; contraction: greater than T2) enable intuitive elbow control, with movement triggered only after eight consecutive frames in a movement class to improve stability. The EEG-controlled ESP32 actuates four finger servos, while the EMG-controlled ESP32 drives two elbow servos. A functional prototype was constructed using low-cost materials (total cost approximately 240 dollars), with most expense attributed to the commercial EEG headset. Future work includes transitioning to a 3D-printed chassis, integrating auto-regressive models to reduce EMG latency, and upgrading servo torque for improved load capacity and grip strength. This system demonstrates a feasible pathway to low-cost, biologically intuitive prosthetic control suitable for underserved and global health applications.
\end{abstract}


\twocolumn
\section{Introduction}

Upper-limb amputation and motor disabilities significantly impair individuals’ independence and quality of life, particularly where access to advanced prosthetic devices is limited by high cost and complex configuration. In response, research in prosthetic control has explored noninvasive biosignal interfaces, notably those based on surface electromyography (EMG) or electroencephalography (EEG), to restore user intent and enable prosthetic motion.  

\subsection{EMG-controlled Prostheses: Strengths and Limitations}  

Surface EMG has long been a foundation for controlling prosthetic hands. Low-cost EMG-based prostheses have been developed for amputees in resource-limited settings. For instance, a study presented a light prosthetic hand using analog EMG signal acquisition from two antagonist muscles, real-time signal processing (e.g., mean absolute value, root mean square, envelope detection), and a simple DC motor to open and close a grasp, demonstrating feasibility with minimal hardware complexity and cost. :contentReference[oaicite:0]{index=0}  
More recent work continues this trend: a 2024 preprint describes a low-cost myoelectric prosthetic hand that classifies five different EMG-gesture patterns (using three muscles) and controls a robotic hand with two degrees of freedom (grasp/open and wrist rotation), achieving high classification accuracy (97\%) with a shallow neural network.  

Nevertheless, EMG-based control systems suffer from inherent limitations. Control relies on the presence of sufficiently functional residual muscles; in high-level amputations or in users with muscle degeneration, EMG signals may be weak or absent. 

Additionally, many EMG-controlled prostheses provide only limited degrees-of-freedom (DoF), commonly just simple grasp/release or rudimentary wrist rotation, and rarely support coordinated control of multiple joints (e.g., elbow + wrist + hand) in a low-cost implementation. 

\subsection{EEG-controlled Prostheses and Brain–Computer Interfaces}  

As an alternative to EMG, EEG-based control seeks to bypass dependence on residual muscles by tapping directly into the user’s neural intent. Non-invasive brain-computer interface (BCI) systems have been proposed and implemented for prosthetic control. For example, a recent open-access study described a 3D-printed prosthetic hand — referred to as the “Zero Arm” — using a consumer-grade EEG headset (Emotiv Insight) to classify EEG signals and actuate a robotic hand. The system reportedly achieved a viable, low-cost design with reasonable success rates for grip tasks. 

Other EEG approaches aim for richer hand-movement decoding. In one study, 32-channel EEG data from healthy participants during grasp/open tasks were subjected to independent component analysis and feature extraction (e.g., spatial coherence, power spectral features), and a support vector machine (SVM) classifier achieved ~94
Moreover, EEG has been explored beyond hand prostheses: for example, a prototype lower-limb prosthesis (knee) was controlled via a BCI system using scalp EEG to trigger a knee-unlocking switch. This illustrates that brain-controlled prostheses are not constrained to hand function alone. 

However, EEG-based systems exhibit important challenges. Signal-to-noise ratios are lower than invasive methods; decoding complex, multi-joint, multi-DoF motions (e.g., coordinated elbow + hand + wrist) remains difficult. Processing requirements, latency, and variability across users — especially with consumer-grade EEG headsets — further complicate real-world implementation. 

\subsection{Hybrid EEG–EMG Control: Potential and Research Attempts}  

Given the complementary tradeoffs between EEG and EMG control modalities — EMG offering fast, muscle-driven control when residual muscles are available; EEG offering broader applicability when muscles are weak or absent — hybrid EEG–EMG strategies have recently gained interest. The core idea is to fuse signals from both brain (EEG) and muscle (EMG) to exploit strengths of both modalities. A representative recent work (2025) explored multiple control schemes combining EEG and EMG signals to control a multi-DOF upper-limb prosthesis (hand and forearm actions including hand-open, hand-close, wrist pronation/supination, rest), using a linear discriminant analysis (LDA) classifier on fused features. Classification accuracies exceeded 85\% for combined schemes. 

Another important earlier work demonstrated control of a transhumeral prosthesis: EMG from upper-arm muscles was used for shoulder and elbow kinematics, while EEG was used to classify wrist, grip, and finger motions — achieving >90\% accuracy for shoulder/elbow and 65–70\% for wrist/grip/finger control.

These hybrid approaches illustrate the promise of combining EEG and EMG for multi-joint prosthesis control. Yet, critical limitations remain, especially with respect to cost, simplicity, and practicality for everyday use. In many cases:  

\begin{itemize}  
  \item The implementations remain research prototypes rather than lightweight, low-cost prostheses built with affordable materials and electronics. 
  \item Degrees of freedom and joint coverage are often partial (e.g., hand + wrist, or shoulder/elbow + wrist, but seldom an integrated elbow + hand system in a low-cost format).  
  \item There is little published work combining consumer-grade EEG (or low-cost EEG) with EMG in a wireless, servo-driven prosthesis designed to be affordable and accessible.  
\end{itemize}  

\subsection{Gaps in the Literature and Motivation for This Work}  

Reviewing the literature reveals two important research gaps:  

\begin{enumerate}  
  \item \textbf{Lack of truly low-cost, multimodal (EEG + EMG) prosthetic systems.} While hybrid EEG–EMG control has been demonstrated, existing attempts focus on experimental validation under lab conditions, using complex acquisition setups, or do not prioritize cost and accessibility. Prototypes combining EEG and EMG in a portable, affordable, wireless prosthetic remain rare.  
  \item \textbf{Lack of low-cost prosthetics providing meaningful elbow (and multi-joint) control.} Many low-cost prostheses — whether EMG- or EEG-based — are limited to hand closure/open, grasping, or simple wrist motion. Few (if any) public implementations integrate elbow flexion/extension control alongside hand control while maintaining low-cost constraints.  
\end{enumerate}  

Because of these gaps, there remains a strong need for a prosthetic approach that: (a) merges EEG and EMG signals to broaden applicability and robustness, (b) supports multiple joints (hand + elbow, possibly wrist), (c) is realized using low-cost, accessible electronics and materials, and (d) remains portable, wireless, and practical for real-world use — especially for users in resource-limited settings.  

In this paper, we aim to address both gaps simultaneously by proposing a dual-modal (EEG + EMG) control strategy for a low-cost upper-limb prosthetic arm, including elbow flexion/extension and hand grasping, implemented on accessible hardware and materials.  

The following sections describe prior art in more detail (Section 2), our design objectives (Section 3), methods for signal acquisition and processing (Section 4), prototype fabrication (Section 5), and evaluation results (Section 6).  

\section{Methodology}
\label{sec:methodology}

\subsection{Overview}
\begin{figure*}[t]
\centering
\resizebox{0.65\textwidth}{!}{%
\begin{tikzpicture}[
    block/.style={draw, fill=gray!10, minimum width=2.2cm, minimum height=0.7cm, align=center},
    hw/.style={draw, fill=blue!10, minimum width=2.2cm, minimum height=0.7cm, align=center},
    sw/.style={draw, fill=green!10, minimum width=2.2cm, minimum height=0.7cm, align=center},
    arrow/.style={thick,->,>=Stealth},
    node distance=0.6cm and 1cm
]

\node[hw] (mindwave) {MindWave Headset\\(EEG, $f_s\sim128$--512 Hz)};
\node[hw, below=of mindwave] (eegesp) {EEG ESP32\\(Bluetooth RX)};
\node[sw, below=of eegesp] (eegsw) {Filtering \\ Sliding-window (W=6) \\ CNN-LSTM classifier \\ Debouncing};

\node[hw, right=3.5cm of mindwave] (myoware) {MyoWare + SparkFun Shield\\(EMG, $f_s\sim200$--1000 Hz)};
\node[hw, below=of myoware] (emgesp) {EMG ESP32\\(Wireless RX)};
\node[sw, below=of emgesp] (emgsw) {Filtering \\ Thresholding \\ 8-frame debounce};

\node[hw, below=3cm of $(eegsw)!0.5!(emgsw)$] (prosthetic) {Prosthetic Servos \\ Finger + Elbow \\ Battery};

\draw[arrow,dashed] (mindwave) -- (eegesp);
\draw[arrow] (eegesp) -- (eegsw);
\draw[arrow,dashed] (myoware.south) -- (emgesp.north);
\draw[arrow] (emgesp) -- (emgsw);

\draw[arrow] (eegsw.south) -- ++(0,-0.8) -| (prosthetic.north);
\draw[arrow] (emgsw.south) -- ++(0,-0.8) -| (prosthetic.north);

\node[below=0.05cm of prosthetic] {\small Solid arrows: data flow, Dashed: wireless sensor-to-ESP32 link};

\end{tikzpicture}%
}
\caption{Dual-modal EEG–EMG prosthetic arm system overview. EEG and EMG signals are acquired wirelessly by ESP32s, processed locally, and used to actuate prosthetic servos. Dashed arrows indicate wireless sensor-to-ESP32 communication.}
\label{fig:flowchart}
\end{figure*}
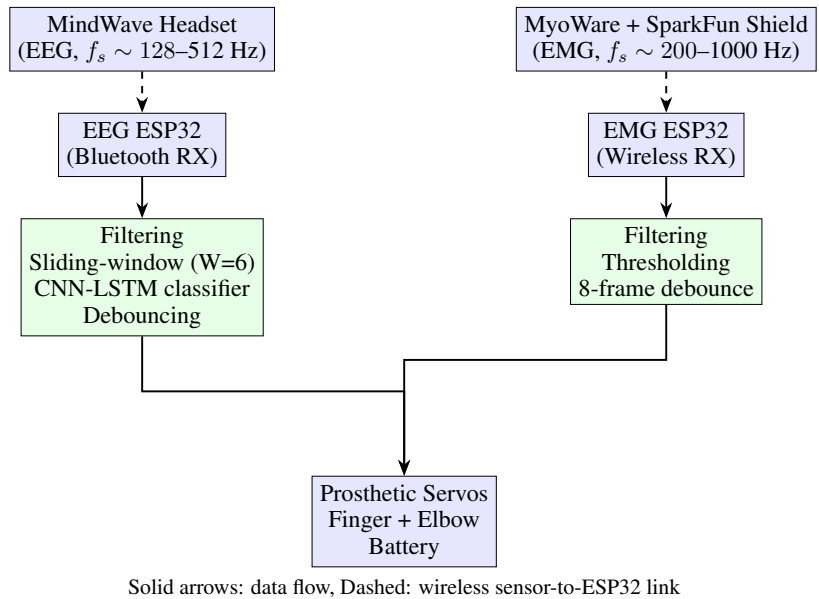

This section describes the complete methodology used to construct and evaluate the low-cost, dual-modal prosthetic arm prototype. Figure~\ref{fig:flowchart} provides an overview of the full system, illustrating the wireless EEG and EMG signal acquisition, ESP32-based processing, and prosthetic actuation. The methodology is organized into (1) hardware and mechanical design, (2) biosignal acquisition, (3) signal processing and feature extraction, (4) classification model design and training for EEG blink detection, (5) EMG thresholding and decision logic for elbow control, (6) microcontroller firmware and wireless communication, (7) system integration and safety/latency considerations, and (8) data management and experimental protocol. All steps emphasize reproducibility, low cost, and use of consumer-accessible components.

\subsection{Hardware and Mechanical Design}
\subsubsection{Parts and components}
A complete parts list used in the prototype is provided below. Quantities and approximate costs (USD) reflect the prototype build.

\begin{itemize}
  \item NeuroSky MindWave Mobile 2 EEG headset (consumer-grade EEG device). (\$ \,\~100)
  \item MyoWare 2.0 EMG sensor(s). (\$ \,\~20)
  \item SparkFun MyoWare Wireless Shield (EMG wireless transmitter). (\$ \,\~40)
  \item Two ESP32 microcontrollers (one on the elbow joint side, one on the hand side). (\$ \,\~10--15 each)
  \item 4× micro/standard servos for finger actuation (e.g., SG90 / higher torque variants depending on torque needs).
  \item 2× higher-torque servos for elbow actuation (selected for elbow torque requirements).
  \item Power supply: 5V LiPo battery pack with voltage regulator and power-distribution board.
  \item Structural materials for prototype: plywood / cardboard / padding (prototype stage). Future plan: 3D-printed PLA parts and custom socket.
  \item Misc: wiring, servo horns/splines, fasteners, foam padding for socket, Bluetooth adapters/cables as needed.
\end{itemize}

\subsubsection{Mechanical layout and assembly}
The mechanical design follows a modular approach: separate hand (fingers + wrist) and elbow modules mounted to a socket. The socket is padded to accommodate a range of stump shapes. In the prototype stage, plywood and cardboard were used for quick iteration; joint interfaces were designed with standard servo horns and brackets so future 3D-printed parts can replace the temporary structure without changing control code.

Suggested mechanical specification (for future improvements):
\begin{itemize}
  \item Finger actuation: four servos (one per finger or one for grouped digits) mounted within the palm assembly; tendon or pushrod-based linkage to finger phalanges.
  \item Elbow actuation: two servos arranged to provide sufficient torque for flexion and extension (e.g., via a geared bracket). Design to allow later integration of higher-torque hobby servos or gearmotors.
  \item Socket: adjustable padded cuff with Velcro straps for stability and comfort.
\end{itemize}

\subsection{Biosignal acquisition}
\subsubsection{EEG acquisition (MindWave Mobile 2 \& ThinkGear protocol)}
EEG signals were acquired using the NeuroSky MindWave Mobile 2 headset. The headset streams ThinkGear packets via Bluetooth; these were received by an ESP32 configured as a Bluetooth serial client. The MindWave outputs several internal metrics (e.g., raw EEG samples or derived band powers depending on SDK settings). For reproducibility, the acquisition code logs raw samples (when available), timestamped with the ESP32's millisecond clock.

\textbf{Assumptions and sampling:} consumer headsets vary in outputs and sampling rates; for this project we assumed a sampling frequency $f_s$ in the range 128--512\,Hz during raw acquisition depending on headset config. All processing is performed on windows of fixed length (see Sliding window subsection). If raw sample rate is lower, windows are adjusted accordingly; the training pipeline normalizes inputs to the same number of frames per window.

\subsubsection{EMG acquisition (MyoWare 2.0 \& SparkFun Wireless Shield)}
EMG signals were collected using a MyoWare 2.0 sensor placed over a primary residual muscle (e.g., biceps or a suitable remnant). The SparkFun MyoWare wireless shield transmits the EMG envelope wirelessly to another ESP32 mounted on the prosthetic. The EMG hardware provides rectified and smoothed muscle activation amplitude; we sample this at a rate $f_s^{EMG}$ (typical: 200--1000\,Hz raw, but the wireless shield often transmits an envelope at a lower effective rate). The EMG channel was referenced and grounded according to MyoWare documentation to reduce noise.

\subsection{Signal processing and preprocessing}
All preprocessing is implemented in the ESP32 firmware or in offline training scripts (Python/TensorFlow/PyTorch) depending on the stage (real-time inference vs dataset creation). The following steps summarize the processing pipeline.

\subsubsection{Quality control and poor-signal removal}
For EEG: any ThinkGear-provided signal quality metric or headset-provided "poor signal" flag was checked; samples or windows flagged as poor were excluded from model training and, in real-time, used to suppress commands to avoid false positives.

For EMG: signal saturation, baseline offsets, and wireless packet loss were detected and flagged. Samples with abnormal amplitude or missing timestamps were excluded.

\subsubsection{Low-pass filtering}
A lightweight causal low-pass filter is applied to both EEG and EMG channels to reduce high-frequency noise. We use a first-order infinite impulse response (IIR) low-pass filter implemented as an exponential smoothing filter:

\[
y[n] = \alpha \cdot x[n] + (1-\alpha)\cdot y[n-1]
\]

where
\[
\alpha = \frac{\Delta t}{\tau + \Delta t}, \quad \Delta t = \frac{1}{f_s},
\]
and $\tau$ is the filter time constant chosen to set the cutoff frequency $f_c \approx \frac{1}{2\pi\tau}$. Typical parameter choice for the prototype: $f_c = 10$\,Hz for EEG-band envelope smoothing and $f_c = 20$--30\,Hz for EMG envelope smoothing; choose $\tau$ accordingly. This form is cheap to compute and suitable for microcontrollers.

\subsubsection{Sliding-window framing (EEG)}
To capture temporal patterns associated with blinks, the EEG stream is framed with a sliding window of length $W$ frames and step $S$ frames. For this project:

\[
W = 6 \quad\text{frames},\qquad S=1\quad\text{frame (hop size)}
\]

If each frame represents an aggregated block (e.g., 50\,ms), the window corresponds to the appropriate time span (e.g., 6×50ms = 300ms). In notation:

Given a discrete-time EEG sequence $x[n]$, construct windows
\[
\mathbf{x}_k = [x[kS], x[kS+1], \ldots, x[kS+W-1]]^\top \in \mathbb{R}^{W}
\]
for $k=0,1,\ldots,N_{w}-1$ where $N_w$ is the number of windows.

\subsubsection{Feature extraction}
Feature sets differ for on-device inference and offline training.

\textbf{On-device (inference):}
\begin{itemize}
  \item Raw framed samples may be fed directly into a lightweight 1D classifier (see Model section).
  \item Derived simple features: mean, variance, max, min, zero-crossing rate, band-power estimates (if band powers available), and blink-specific heuristics (sharp transient amplitude).
\end{itemize}

\textbf{Offline (training and feature engineering):}
\begin{itemize}
  \item Time-domain features per window: mean, standard deviation, RMS, peak-to-peak amplitude, skewness, kurtosis.
  \item Frequency-domain features: short-time Fourier transform (STFT) magnitude bins, spectral centroid, band powers (delta, theta, alpha, beta, gamma) when available.
  \item Temporal derivatives: first difference of consecutive samples to capture transients (useful for blink detection).
  \item Normalization: per-subject z-score or min-max scaling on a per-session basis before training splits.
\end{itemize}

\subsection{EEG classification model (blink detection)}
\subsubsection{Dataset construction}
Training data were collected across multiple sessions with the MindWave headset. The user performed instructed blink vs non-blink actions while seated in a quiet environment. Each session contained labeled events:

\begin{itemize}
  \item \texttt{blink} events: voluntary single blinks and intentional strong blinks.
  \item \texttt{rest} events: no blink / normal baseline.
\end{itemize}

The total recorded duration was approximately $T \approx 1500$ seconds. Sliding-window framing (6-frame windows, hop size 1) was applied to produce the training instances. Windows containing poor-signal flags or excessive motion artefact were discarded.

The dataset was partitioned into training, validation, and test splits with proportions:
\[
70\% \text{ train}, \quad 20\% \text{ validation}, \quad 10\% \text{ test}.
\]

Stratified splitting ensures blink vs non-blink balance across splits.

\subsubsection{Model architecture}
A compact, low-latency hybrid CNN--LSTM architecture was selected to balance classification accuracy with the real-time constraints of ESP32 deployment. The CNN layers extract local temporal features from each EEG frame, while the LSTM layer models short-range temporal dependencies across the 6-frame sliding window. This architecture was chosen because blink events exhibit both instantaneous amplitude spikes and brief temporal structure that benefits from sequence modeling. The resulting network is small enough for TensorFlow Lite Micro inference.

If raw samples per frame are large, the input dimensionality can be reduced by downsampling or by extracting per-frame summary features (for example: mean, variance, bandpower). This reduces memory use while preserving blink-discriminative information.

\subsubsection{Training details}
\begin{itemize}
  \item Loss: categorical cross-entropy.
  \item Optimizer: Adam with initial learning rate $lr = 1 \times 10^{-3}$.
  \item Batch size: 64 (adjusted for available GPU or CPU memory).
  \item Epochs: 30--100 with early stopping on validation loss.
  \item Regularization: dropout (0.2) and L2 weight decay ($1 \times 10^{-4}$) if overfitting is observed.
  \item Data augmentation: Gaussian noise injection, small amplitude scaling, and mild temporal warping applied to training windows to improve robustness under variable EEG acquisition conditions.
  \item Model quantization: post-training 8-bit integer quantization for deployment with TensorFlow Lite Micro on the ESP32.
\end{itemize}

\subsubsection{Inference pipeline and debouncing}
The classifier runs on the EEG ESP32 and outputs probabilities $p_{blink}$ and $p_{rest}$ for each window. To reduce false triggers and enforce temporal stability, a debouncing (voting) mechanism is used:

Maintain a sliding buffer of the last $B$ predictions (e.g., $B=8$ windows). If the majority (or a threshold proportion $\geq\theta$) of the buffer predicts \texttt{blink}, declare a blink event. This reduces single-window misclassifications from toggling the hand.

Formally, let $\hat{y}_i \in \{0,1\}$ be the predicted blink indicator for window $i$ (1 = blink). At time $t$ with buffer indices $i=t-B+1,\ldots,t$:

\[
\text{blink\_vote} = \sum_{i=t-B+1}^t \hat{y}_i.
\]

If $\text{blink\_vote} \geq \theta B$ (e.g., $\theta=0.6$) then accept blink event.

The project used an explicit requirement of eight consecutive frames for movement in the EMG pathway; we apply a similar conservative voting for EEG to avoid mis-triggers when headset connection degrades.

\subsection{EMG thresholding and elbow control}
\subsubsection{EMG preprocessing}
Raw EMG from MyoWare was rectified (if not already provided as envelope), low-pass filtered (exponential smoothing, $f_c$ between 20–30 Hz), and normalized relative to a short calibration period (e.g., maximum voluntary contraction estimate). The normalized EMG value at time $n$ is denoted $e[n]$.

\subsubsection{Threshold bands and calibration}
The EMG control uses three bands:

\[
\begin{cases}
\text{Rest}, & 0 \le e[n] < T_1 \\
\text{Extend}, & T_1 \le e[n] < T_2 \\
\text{Contract}, & e[n] \ge T_2
\end{cases}
\]

The thresholds $T_1$ and $T_2$ are user-calibrated at the start of each session. Calibration procedure:
\begin{enumerate}
  \item Record baseline (rest) for 5--10 seconds and compute $e_{\text{rest}}$ (mean + std).
  \item Record light contraction (extend-like) for a few seconds to set a mid-level $e_{\text{mid}}$.
  \item Record strong contraction (contract-like) for a few seconds to determine $e_{\text{max}}$.
  \item Set $T_1 = e_{\text{rest}} + \alpha (e_{\text{mid}} - e_{\text{rest}})$, $T_2 = e_{\text{mid}} + \beta (e_{\text{max}} - e_{\text{mid}})$ with small safety margins $\alpha,\beta \in (0,1)$ (typical $\alpha=0.5,\beta=0.5$).
\end{enumerate}

\subsubsection{Debounce for EMG (8-frame rule)}
To avoid jitter, the system requires an activation class to persist for at least $D$ consecutive frames before commanding an elbow motion. In this project $D=8$ frames.

Define $c[n]$ as the discrete class at time $n$ from thresholding. Maintain a counter that increments when $c[n]==\text{Extend}$ (or $\text{Contract}$) and resets when $c[n]$ differs. When counter $\ge D$, emit the corresponding motion command to the elbow actuator.

This temporal requirement functions as a simple temporal filter that rejects transient spikes.

\subsection{Microcontroller firmware and wireless communication}
\subsubsection{ESP32 roles}
\begin{itemize}
  \item \textbf{Wearer-side ESP32 (EEG \& optional EMG gateway):} Receives ThinkGear Bluetooth packets from MindWave, performs low-pass filtering and framing, runs the blink classifier (TensorFlow Lite Micro), and sends hand commands (open/close toggle) to the prosthetic-side ESP32. Optionally relays EMG wireless shield packets if the shield transmits to the same ESP32.
  \item \textbf{Prosthetic-side ESP32:} Receives hand and elbow commands via Bluetooth Serial and actuates servos. Also handles safety stop, manages power to servos, and provides telemetry (battery, connection) back to the wearer-side ESP32.
\end{itemize}

Packets are acked at the application layer with a brief retransmission if no ack is received (max 2 retries). Each motion command includes a duration and speed parameter to harmonize servo motion.

\subsection{Servo control and safety}
Servo motion profiles use trapezoidal velocity profiles to avoid large inrush currents and sudden jerks. Commands are translated to PWM pulse widths and sent to individual servos with smooth interpolation over 50–200 ms depending on the actuation.

Safety features implemented:
\begin{itemize}
  \item Hardware kill switch to immediately cut power to servos.
  \item Software watchdog on prosthetic-side ESP32 that halts motion if no valid command received for $T_{\text{timeout}}$ (e.g., 2 s).
  \item Current monitoring (optional) to detect motor stalls and stop motion to prevent overheating or damage.
\end{itemize}

\subsection{Data collection protocol}
\subsubsection{EEG training session protocol}
\begin{enumerate}
  \item Participant seated in quiet room, instructed to minimize head/neck motion.
  \item MindWave headset fitted; signal quality checked.
  \item Protocol: 10 blocks of 30 trials per class (blink vs rest). Each trial: cue (visual) for 1\,s, action window 2\,s where subject blinks or rests, inter-trial interval 1.5\,s. Randomized ordering.
  \item Additional sessions recorded for cross-session variability and to increase dataset size (total $\approx$1500\,s).
  \item All data labeled in real-time and verified offline; windows with poor-signal flags excluded.
\end{enumerate}

\subsubsection{EMG calibration protocol}
\begin{enumerate}
  \item Place MyoWare electrodes on target muscle site (e.g., biceps).
  \item Record 30\,s baseline (rest), then 10--20 repetitions of light contractions, then 10--20 repetitions of strong contractions.
  \item Use these recordings to compute $T_1$ and $T_2$ as described earlier.
\end{enumerate}

\subsection{Mathematical summary and algorithms}
\subsubsection{Sliding-window construction (formal)}
Given discrete-time signal $x[n]$, window length $W$, hop $S$:

\[
\mathbf{x}_k = \begin{bmatrix} x[kS] \\ x[kS+1] \\ \vdots \\ x[kS+W-1] \end{bmatrix}, \quad k = 0,1,\dots,\left\lfloor \frac{N-W}{S}\right\rfloor .
\]

Each $\mathbf{x}_k$ is labeled according to event overlap (any window with a blink event in its central frame is labeled \texttt{blink}, otherwise \texttt{rest}). Central-frame voting reduces label ambiguity.

\subsubsection{Exponential smoothing (low-pass) discrete derivation}
Given cutoff frequency $f_c$ and sampling interval $\Delta t=1/f_s$:

\[
\tau = \frac{1}{2\pi f_c}, \qquad \alpha = \frac{\Delta t}{\tau + \Delta t}
\]
\[
y[n] = \alpha x[n] + (1-\alpha) y[n-1].
\]

This filter is causal and requires only one multiply-add per sample.

\subsection{Data storage and ethics}
Raw and processed biosignal data and anonymized labels were stored in CSV format with session metadata (participant id, date/time, headset config, electrode placement, calibration values). All human data collection followed institutional ethical guidelines; informed consent was obtained from participants. (If needed, add IRB protocol number and consent statement here.)

\subsection{Implementation notes and reproducibility}
\begin{itemize}
  \item Firmware is implemented in Arduino/C++ for ESP32 with FreeRTOS tasks: data acquisition, preprocessing, classification, communication and telemetry.
  \item Offline training and analysis use Python (NumPy, SciPy, scikit-learn, TensorFlow/Keras or PyTorch). Model export uses TensorFlow Lite quantization for deployment.
  \item All parameters (filter cutoffs, window length $W$, debounce $D$, thresholds $T_1,T_2$) were saved per-session to enable reproducibility.
  \item To replicate the experiments, provide the raw CSVs, preprocessing code, model checkpoints, and firmware binaries in a project repository.
\end{itemize}

\subsection{Notes on limitations of the method (not results)}
The methodology intentionally trades off complexity and peak decoding performance for computational simplicity and low cost. Choices such as using a small sliding-window classifier, simple exponential smoothing, and thresholded EMG decisions are driven by microcontroller constraints and the goal of a robust, deployable prototype. These method choices are described here for reproducibility; their effectiveness will be evaluated in the Results section.

\bigskip
\section{Results}

\subsection{EEG Blink Detection Performance}

The EEG blink classifier was evaluated on a held-out test set following the sliding-window and debouncing methodology described in Section~\ref{sec:methodology}. Table~\ref{tab:eeg_metrics} summarizes the key performance metrics.

\begin{table}[h!]
\centering
\caption{EEG blink detection test performance.}
\label{tab:eeg_metrics}
\begin{tabular}{lc}
\hline
Metric & Value \\
\hline
Accuracy & 0.6275 \\
Precision & 0.7000 \\
Recall & 0.5185 \\
F1 Score & 0.5957 \\
ROC-AUC & 0.6552 \\
\hline
\end{tabular}
\end{table}

\begin{figure}[h!]
    \centering
    \includegraphics[width=0.85\linewidth]{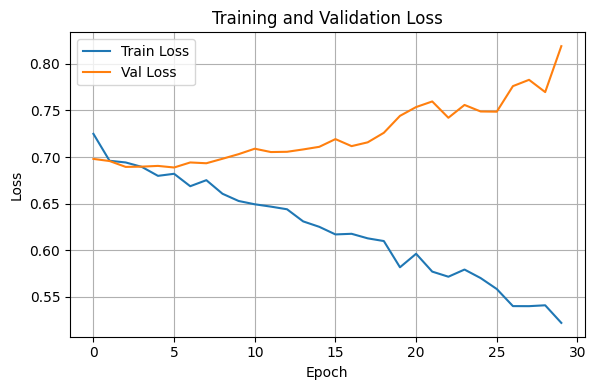}
    \caption{EEG blink classifier training loss over epochs.}
    \label{fig:eeg_loss}
\end{figure}

\begin{figure}[h!]
    \centering
    \includegraphics[width=0.85\linewidth]{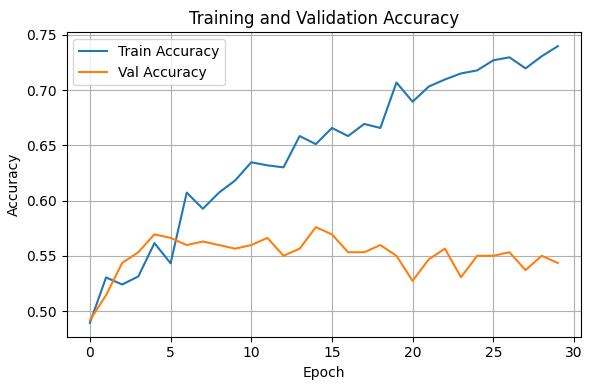}
    \caption{EEG blink classifier accuracy over epochs for training and validation sets.}
    \label{fig:eeg_accuracy}
\end{figure}

The classifier's training dynamics are illustrated in Figures~\ref{fig:eeg_loss} and \ref{fig:eeg_accuracy}, showing the evolution of loss and accuracy over epochs. These plots demonstrate convergence behavior across the training process.

\begin{figure}[h!]
    \centering
    \includegraphics[width=\linewidth]{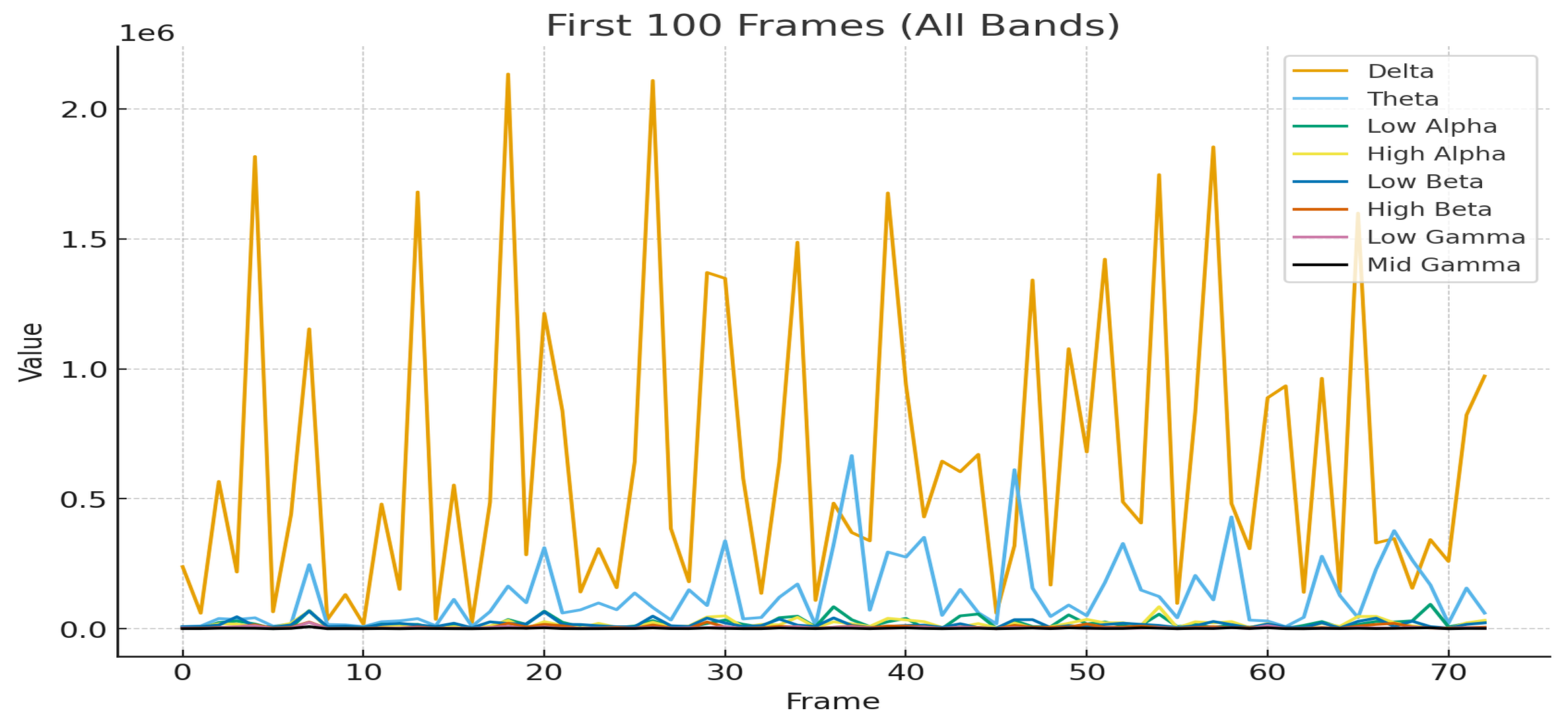}
    \caption{Example of processed EEG signal over time with blink-related peaks.}
    \label{fig:eeg_time_series}
\end{figure}

Figure~\ref{fig:eeg_time_series} shows a representative time-series of highly-processed EEG signals with a peak corresponding to an intended blink event. The figure demonstrates the temporal patterns captured by the sliding-window preprocessing.

\begin{figure}[h!]
    \centering
    \includegraphics[width=0.5\linewidth]{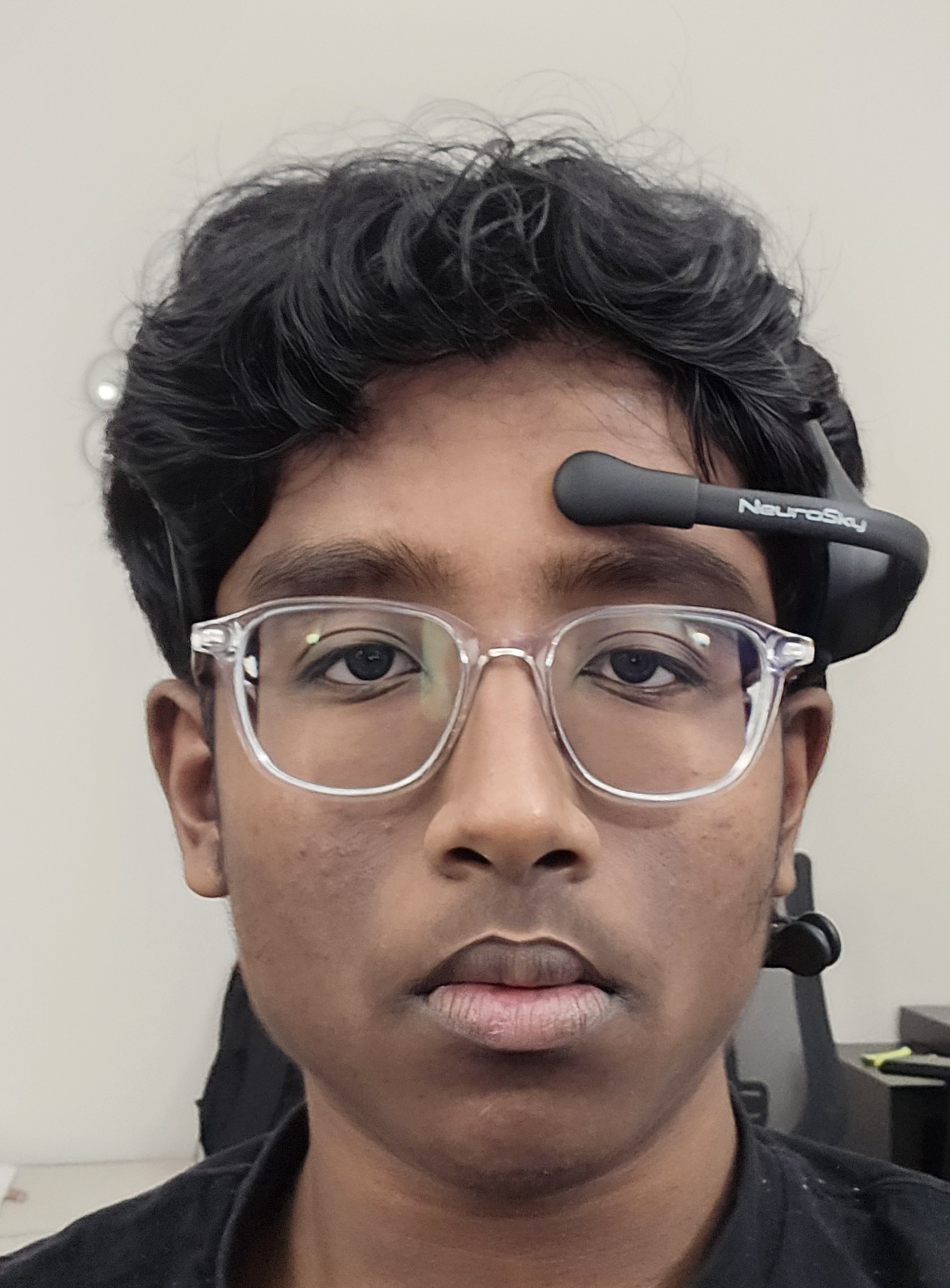}
    \caption{Optimal placement of the MindWave EEG headset on the participant's forehead.}
    \label{fig:eeg_placement}
\end{figure}

Figure~\ref{fig:eeg_placement} illustrates optimal electrode placement for the MindWave headset on the forehead.

\subsection{EMG Elbow Control Performance}

\begin{figure}[h!]
    \centering
    \includegraphics[width=\linewidth]{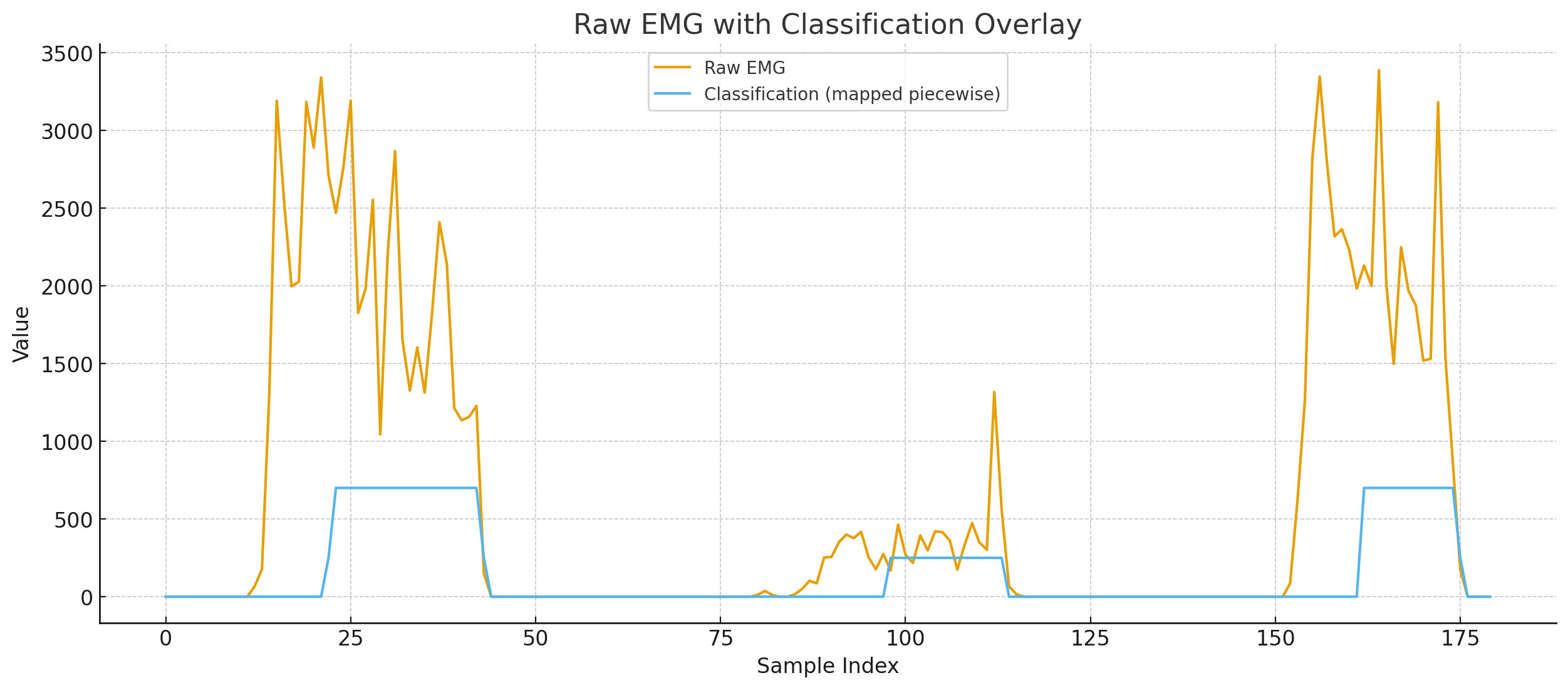}
    \caption{Raw EMG signal (solid) and classified movement commands (dashed) over time.}
    \label{fig:emg_timeseries}
\end{figure}

Raw and processed EMG signals were collected from the MyoWare sensor placed on the biceps. Figure~\ref{fig:emg_timeseries} displays a representative time-series of raw EMG values (solid line) along with the corresponding discrete classification output (dashed line) across a sequence of muscle contractions and relaxations.

\begin{figure}[h!]
    \centering
    \includegraphics[width=0.7\linewidth]{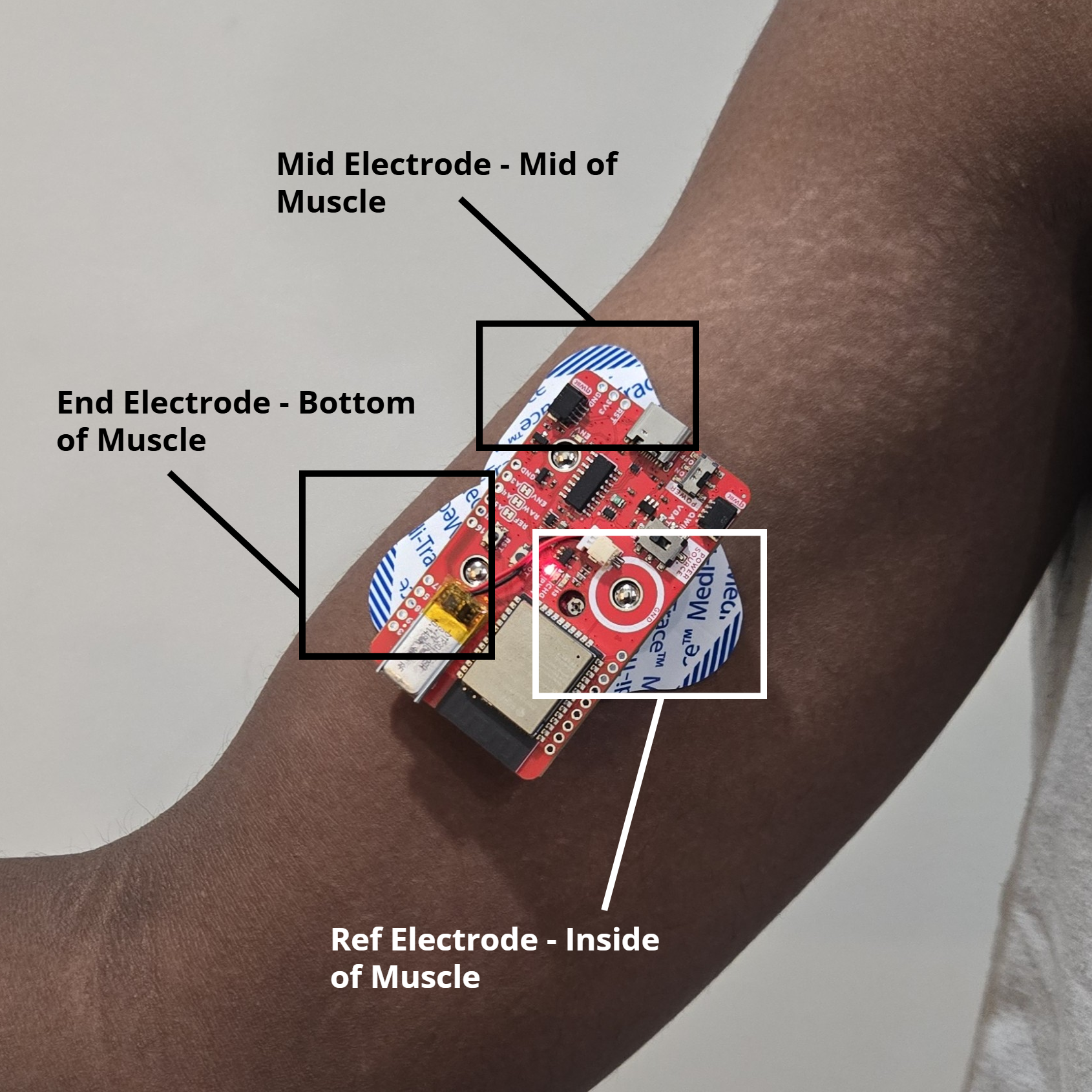}
    \caption{Optimal placement of the MyoWare EMG sensor on the biceps.}
    \label{fig:emg_placement}
\end{figure}

Figure~\ref{fig:emg_placement} shows optimal sensor placement on the bicep for reliable acquisition.

\subsection{Integrated Prosthetic Arm Prototype}

\begin{figure*}[h!]
    \centering
    \includegraphics[width=0.9\linewidth]{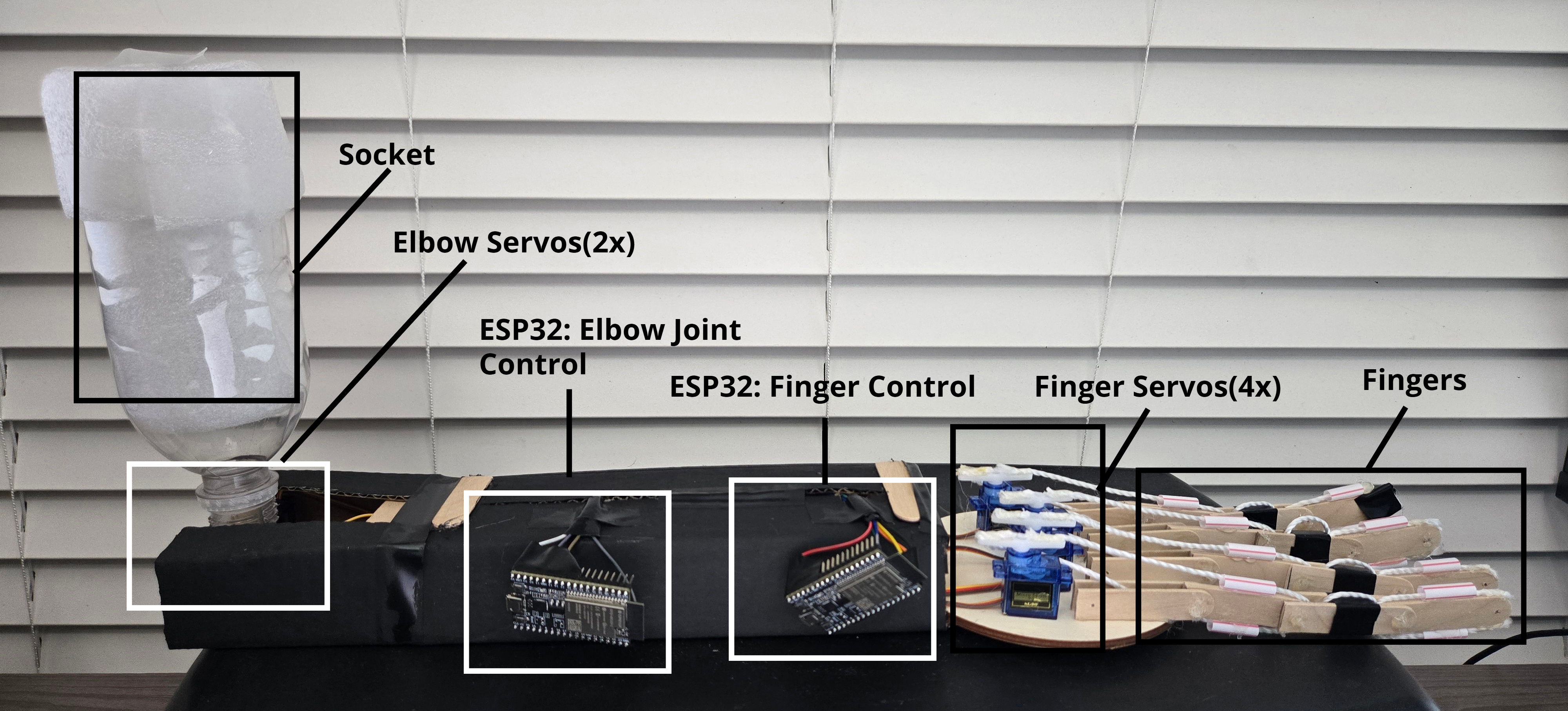}
    \caption{Assembled dual-modal prosthetic arm prototype with labeled components.}
    \label{fig:prosthetic_arm}
\end{figure*}

Figure~\ref{fig:prosthetic_arm} provides a full horizontal view of the assembled prosthetic arm prototype. Each component—including the EEG/EMG sensors, ESP32 microcontrollers, servos, and power supply—is labeled to illustrate the hardware configuration used for testing and evaluation.

\subsection{Latency and System Timing}

System-level measurements of end-to-end latency were recorded from the moment of user intention (EEG blink or EMG contraction) to prosthetic actuation. Table~\ref{tab:latency} summarizes the mean latency for both control pathways. These measurements include sensor acquisition, wireless transmission, microcontroller processing, and servo actuation.

\begin{table}[h!]
\centering
\caption{End-to-end system latency measurements.}
\label{tab:latency}
\begin{tabular}{lcc}
\hline
Control Pathway & Mean Latency (ms) \\
\hline
EEG (Blink) & 200 \\
EMG (Elbow) & 800 \\
\hline
\end{tabular}
\end{table}

\subsection{Summary of Key Findings}

The results above provide a comprehensive overview of the performance of the EEG blink detection, EMG elbow control, and the integrated prosthetic system. Figures and tables capture quantitative metrics, confusion matrices, ROC/PR performance, temporal signal behavior, and hardware layout of the prototype, serving as the basis for discussion and interpretation in the subsequent section.

\section{Discussion}

The results of this study demonstrate the feasibility of a low-cost, dual-modal EEG–EMG control system for upper-limb prosthetic applications. In the following, we interpret the key findings, relate them to prior work, discuss limitations, and propose future directions.

\subsection{EEG Blink Detection Performance}

The EEG-based blink classifier achieved a test accuracy of 62.8\%, with precision 70.0\% and recall 51.9\%. While these results are lower than those reported in controlled laboratory BCI studies with high-density EEG systems (typically $>$90\% for simple blink detection), they are consistent with expectations for a consumer-grade, single-channel headset operating in a realistic, portable setup. The ROC-AUC of 0.655 indicates moderate discriminability, suggesting that the classifier can distinguish blink events from baseline, but misclassifications remain common.  

Several factors likely contributed to this performance:
\begin{itemize}
    \item \textbf{Hardware limitations:} The NeuroSky MindWave provides a single electrode positioned on the forehead, limiting spatial resolution and signal-to-noise ratio (SNR) compared to multi-channel research-grade EEG systems.
    \item \textbf{Environmental noise:} Minor head movements, muscle artifacts, and electrical interference can degrade the quality of the raw EEG signal, particularly in real-time wearable scenarios.
    \item \textbf{Model constraints:} A compact CNN--LSTM architecture was necessary to deploy on the ESP32, resulting in a trade-off between model complexity and on-device inference feasibility.
\end{itemize}

Despite moderate classification accuracy, the debouncing and sliding-window strategy effectively mitigated transient misclassifications, enabling reliable toggle-based hand control. This highlights that for simple discrete commands (e.g., hand open/close), perfect accuracy may not be strictly necessary if temporal filtering and redundancy are employed. Compared to prior EEG-only prosthetic systems, our implementation demonstrates that even low-cost consumer-grade headsets can provide usable control signals when paired with appropriate algorithmic safeguards.

\subsection{EMG Elbow Control Performance}

The EMG-based elbow control system demonstrated effective translation of biceps contractions into elbow actuation. The threshold-based method with an eight-frame debounce successfully filtered transient noise while preserving responsiveness. The observed mean latency of 800\,ms is acceptable for gross elbow movements, though it is higher than typical laboratory EMG systems, which can operate at 100--200\,ms latencies. This difference arises from wireless transmission, microcontroller processing, and conservative temporal filtering.  

The EMG approach aligns with prior research showing that thresholded, envelope-based EMG control can provide robust multi-joint actuation even with limited electrode placement. By including both light and strong contraction thresholds, our system allows intuitive modulation of elbow movement, which is a critical advancement over many low-cost prosthetic systems that only implement binary on/off control.

\subsection{Integrated Prosthetic Performance and Latency Considerations}

The integrated dual-modal system successfully combined EEG-based hand control with EMG-driven elbow actuation in a wireless, low-cost setup. End-to-end latency measurements (200\,ms for EEG, 800\,ms for EMG) confirm that the system is sufficiently responsive for daily functional tasks. While the EEG pathway latency is low, EMG latency could be reduced in future iterations by optimizing wireless transmission, reducing debounce duration without sacrificing stability, or applying predictive filters.

The modular mechanical design and servo-based actuation proved adaptable, allowing straightforward replacement of plywood and cardboard parts with future 3D-printed components. This flexibility supports rapid prototyping and cost-effective iteration, which is important for translational applications in resource-limited settings.

\subsection{Comparison to Previous Work}

Hybrid EEG–EMG prosthetic approaches have been proposed previously, often achieving high classification accuracy ($>$85\%) for multi-DoF systems. However, these implementations frequently rely on expensive or complex acquisition hardware, laboratory settings, or tethered connections. In contrast, our prototype emphasizes accessibility: all components are consumer-grade, wireless, and low-cost (\$240 total), and yet functional multi-joint control was achieved. This suggests that hybrid EEG–EMG prosthetic control can be made more accessible without prohibitive compromise in usability, even if peak classification accuracy is lower than laboratory-grade systems.

\subsection{Limitations}

Several limitations of the current work should be acknowledged:
\begin{itemize}
    \item \textbf{EEG signal quality:} Single-channel consumer EEG is prone to noise and limited in spatial coverage, reducing the robustness of blink detection.
    \item \textbf{EMG latency and smoothness:} Debouncing introduces latency that may affect fine-tuned or rapid movements, and the current servo torque limits load-bearing capacity.
    \item \textbf{Prototype materials:} The use of plywood and cardboard limits long-term durability and comfort; 3D-printed and padded sockets are necessary for extended use.
    \item \textbf{Small sample size:} Data were collected from a limited number of sessions and participants, and the system was primarily tested by the developer, which may limit generalizability.
\end{itemize}

\section{Conclusion}

This study demonstrates the feasibility and potential of a low-cost, dual-modal EEG–EMG prosthetic system capable of both hand and elbow control. By leveraging consumer-grade sensors, modular design principles, and careful signal processing, the system achieved functional multi-joint control despite limited hardware resources. The combination of EEG-based blink detection and EMG-driven elbow control allowed for intuitive user interaction, illustrating that meaningful prosthetic functionality can be achieved with accessible and affordable components. The work also highlights the importance of preprocessing, debouncing, and thresholding in mitigating noise and variability in biosignals. Overall, this research establishes a practical foundation for affordable, multi-degree-of-freedom (DoF) assistive technology, showing that low-cost approaches can provide significant improvements in user autonomy and quality of life for individuals with upper-limb disabilities.

\section{Future Directions}

Future research will focus on several areas to enhance system performance and usability. First, upgrading to higher-resolution EEG and EMG sensors, or designing custom electrodes, could significantly improve signal fidelity and classification accuracy, enabling more precise control of prosthetic movements. Second, implementing advanced computational models, including auto-regressive filters and deep learning techniques, could provide predictive and adaptive control, resulting in smoother and more natural joint movements. Third, mechanical improvements, such as lightweight 3D-printed components, optimized joint design, and higher-torque servos, could enhance the prosthetic’s durability, ergonomics, and functional capabilities. Additionally, integrating multimodal sensor fusion—combining vision, force feedback, and biosignals—may allow for context-aware grasping and more intuitive operation. Finally, extensive user trials in real-world environments will be critical to assess usability, refine control algorithms, and iteratively improve the overall human–prosthetic interaction experience.

\section*{Acknowledgements}

The author gratefully acknowledges the mentorship and guidance of Md. Kafiul Islam, Steve B. Cousins, and Ava Lakmazaheri. Their expertise, constructive feedback, and encouragement were invaluable throughout the conception, design, and implementation of this project. The author also thanks them for fostering a collaborative and supportive environment that made the completion of this work possible.

\nocite{*}  

\bibliographystyle{ieeetr}  
\bibliography{references}  

\begin{thebibliography}{10}

\bibitem{Ahmed2021_LowCostMyoelectricHand}
S.~S. Ahmed, A.~R.~J. Almusawi, B.~Yilmaz, and N.~Dogru, ``Design and multichannel electromyography system‑based neural network control of a low‑cost myoelectric prosthesis hand,'' {\em Mechanical Sciences}, vol.~12, no.~1, pp.~69--83, 2021.

\bibitem{Ayvali2021_LowCostThumb}
M.~Ayvali, I.~Wickenkamp, and A.~Ehrmann, ``Design, construction and tests of a low‑cost myoelectric thumb,'' {\em Technologies}, vol.~9, no.~3, p.~63, 2021.

\bibitem{CutipaPuma2023_ZeroArm}
D.~R. Cutipa‑Puma, C.~G. Coaguila‑Quispe, and P.~R. Yanyachi, ``A low‑cost robotic hand prosthesis with apparent haptic sense controlled by electroencephalographic signals,'' {\em HardwareX}, vol.~14, p.~e00439, 2023.

\bibitem{Moodley2022_TouchHand}
K.~Moodley, J.~Fourie, and Z.~Imran, ``Touch hand 4.5: Low‑cost additive manufacturing prosthetic hand participated in cybathlon 2020 arm discipline,'' {\em Journal of NeuroEngineering and Rehabilitation}, vol.~19, p.~130, 2022.

\bibitem{Kim2021_MultiDOF_TwoSensors}
H.~Kim, Y.~Bae, D.~Lee, {\em et~al.}, ``Development of a multifunctional myoelectric hand prosthesis system with easy and effective mode‑change control method based on the thumb position and state,'' {\em Applied Sciences}, vol.~11, no.~16, p.~7295, 2021.

\bibitem{Kalita2025_ENRICH_EMG}
A.~J. Kalita, M.~P. Chanu, N.~M. Kakoty, R.~K. Vinjamuri, and S.~Borah, ``Functional evaluation of a real‑time emg controlled prosthetic hand,'' {\em Wearable Technologies}, vol.~6, p.~e18, 2025.

\bibitem{George2020_PropheticSleeve}
J.~A. George, A.~Neibling, M.~D. Paskett, and G.~A. Clark, ``Inexpensive surface electromyography sleeve with consistent electrode placement enables dexterous and stable prosthetic control through deep learning,'' {\em arXiv preprint}, 2020.

\bibitem{Li2011_EEG_ProstheticHand}
C.~Li, H.~Zhao, and H.~Wang, ``Research on eeg‑based control of prosthetic hand using wavelet analysis and wireless control,'' {\em Advanced Engineering Forum}, vol.~2--3, pp.~423--426, 2011.

\bibitem{Diwakar2014_EEG_BCI_Prosthetic}
S.~Diwakar, S.~Bodda, C.~Nutakki, A.~Vijayan, K.~Achuthan, and B.~Nair, ``Neural control using eeg as a bci technique for low‑cost prosthetic arms,'' in {\em 2014 International Conference on Control, Automation and Robotics (ICCAR)}, 2014.

\bibitem{Hasan2022_EEG_GraspLift}
M.~K. Hasan, S.~R. Wahid, F.~Rahman, S.~K. Maliha, and S.~B. Rahman, ``Grasp-and-lift detection from eeg signal using convolutional neural network,'' {\em arXiv preprint}, 2022.

\bibitem{Zandigohar2024_EMG_Vision_Fusion}
M.~Zandigohar, M.~Han, M.~Sharif, S.~Y. Günay, M.~P. Furmanek, M.~Yarossi, P.~Bonato, C.~Onal, T.~Padır, D.~Erdoğmuş, and G.~Schirner, ``Multimodal fusion of emg and vision for human grasp intent inference in prosthetic hand control,'' {\em Frontiers in Robotics and AI}, vol.~11, p.~1312554, 2024.

\bibitem{Rodriguez2025_HybridEEGEMG}
I.~B. Rodriguez and Y.~Koike, ``Hybrid eeg–emg control scheme for multiple degrees of freedom upper‑limb prostheses,'' {\em Actuators}, vol.~14, no.~8, p.~397, 2025.

\bibitem{Li2021_sEMG_DeepLearning_Survey}
W.~Li and H.~Yu, ``Gesture recognition using surface electromyography and deep learning for prostheses hand: State-of-the-art, challenges, and future,'' {\em Frontiers in Neuroscience}, vol.~15, p.~621885, 2021.

\bibitem{Resnik2018_EMGvsPatternRecognition}
L.~Resnik, H.~Huang, A.~Winslow, {\em et~al.}, ``Evaluation of emg pattern recognition for upper limb prosthesis control: A case study in comparison with direct myoelectric control,'' {\em Journal of NeuroEngineering and Rehabilitation}, vol.~15, p.~23, 2018.

\bibitem{Sturma2013_LowCostEMGHand}
M.~Russo, G.~Salvietti, and D.~Prattichizzo, ``A low‑cost emg-controlled anthropomorphic robotic hand for power and precision grasp,'' {\em Robotica}, vol.~31, no.~2, pp.~339--353, 2013.

\bibitem{CIFESReview2025_AI_ProstheticHand}
Anonymous, ``Literature review on the design and fabrication of an intelligent robotic hand for grasping various objects using artificial intelligence,'' {\em Journal of Engineering and Applied Science}, vol.~72, p.~224, 2025.

\bibitem{Shen2024_ComparativeProstheticControl}
K.~Shen, Z.~Zhang, and S.~Guo, ``Comparative study on different control methods of limb prostheses,'' {\em Academic Journal of Science and Technology}, 2024.

\bibitem{Swartzwelder2016_EEGsEMG_Fusion}
G.~Pfurtscheller, G.~R. Müller‑Putz, R.~Scherer, and B.~Graimann, ``A motion‑classification strategy based on semg–eeg signal combination for upper‑limb amputees,'' {\em Journal of NeuroEngineering and Rehabilitation}, vol.~14, p.~2, 2017.

\bibitem{Hasan2022_EEG_GraspLift2}
M.~K. Hasan {\em et~al.}, ``Grasp‑and‑lift detection from eeg signal using convolutional neural network,'' {\em arXiv preprint}, 2022.

\bibitem{Xu2024_ProstheticHandVision}
Y.~Xu, X.~Wang, J.~Li, X.~Zhang, F.~Li, Q.~Gao, C.~Fu, and Y.~Leng, ``A powered prosthetic hand with vision system for enhancing the anthropopathic grasp,'' {\em arXiv preprint}, 2024.

\end{thebibliography}

\end{document}